# К ВОПРОСУ РАЗРАБОТКИ ИНСТРУМЕНТАЛЬНОГО КОМПЛЕКСА ОНТОЛОГИЧЕСКОГО НАЗНАЧЕНИЯ

*А.В. Палагин, Н.Г. Петренко, В.Ю. Величко, К.С. Малахов, Ю.Л. Тихонов*

Киев-187 МСП, 03680, проспект Академика Глушкова, 40, Институт кибернетики имени В.М. Глушкова НАН Украины, e-mail: palagin_a@ukr.net; malakhovks@nas.gov.ua, факс: +38 044 5263348

Предлагаются методологические основы проектирования инструментального комплекса онтологического назначения, предназначенного для реализации интегрированной информационной технологии автоматизированного построения онтологий предметных областей. Полученные результаты ориентированы на повышение эффективности анализа и понимания естественно-языковых текстов, построения баз знаний предметных областей (в первую очередь в научно-технической сфере) и проведения междисциплинарных научных исследований в совокупности с решением сложных прикладных задач.

Пропонуються методологічні основи проектування інструментального комплексу онтологічного призначення, призначеного для реалізації інтегрованої інформаційної технології автоматизованої побудови онтологій предметних областей. Одержані результати орієнтовані на підвищення ефективності аналізу та розуміння природномовних текстів, побудови баз знань предметних областей (насамперед в науково-технічній сфері) та проведення міждисциплінарних наукових досліджень у поєднанні з вирішенням складних прикладних завдань.

The given work describes methodological principles of design instrumental complex of ontological purpose. Instrumental complex intends for the implementation of the integrated information technologies automated build of domain ontologies. Results focus on enhancing the effectiveness of the automatic analysis and understanding of natural-language texts, building of knowledge description of subject areas (primarily in the area of science and technology) and for interdisciplinary research in conjunction with the solution of complex problems.

## Введение

Уровень развития интеллектуальных информационных технологий в значительной мере влияет на эффективность процессов, происходящих в экономической, научно-технической, образовательной и других сферах деятельности человеческого общества. Процессы глобальной информатизации мирового сообщества ориентированы, прежде всего, на построение знание-ориентированного общества и носят все более ярко выраженный междисциплинарный и трансдисциплинарный характер. Несомненным лидером при этом являются технологии инженерии знаний, в том числе ее новое направление – онтологический инжиниринг. Эти технологии реализуют процессы управления знаниями (Knowledge Management) и успехи в этом направлении во многом определяются уровнем интеллектуализации и эффективности компьютерных систем [1–6]. В настоящее время наметилась тенденция к активизации научных исследований на стыке разных предметных дисциплин, в том числе построение знание-ориентированных информационных систем, разработка методов онтологического анализа естественно-языковых объектов (ЕЯО) с целью извлечения из них знаний, прикладных аспектов применения онтологий, в частности при разработке электронных учебных курсов, метаонтологий, систем исследования интеграции знаний в трансдисциплинарных областях и др.

Актуальность указанной тематики на современном этапе развития Computer Science обосновывается следующим образом.

**Во-первых**, большая часть знаний, накопленных за все время развития человеческого общества, содержится в ЕЯО, представленных в текстовой форме. Кроме того, ежедневный прирост новой текстовой информации (где явным лидером является сеть Интернет) привел к тому, что полностью обработать эти текстовые источники знаний, а также получить релевантную запросу пользователя информацию, в компьютерных системах (КС) существующими методами стало практически невозможным. Это приводит к безвозвратной потере части новой информации или ее получение со значительным опозданием, когда она может потерять свою актуальность. Таким образом, задача разработки методологии, принципов и механизмов построения КС с обработкой предметно-ориентированных знаний, представленных в форме естественно-языковых текстов, и на их основе *создания инструментальных средств автоматизированного построения баз знаний предметных областей* (ПдО), имеет важное теоретическое и прикладное значение.

**Во-вторых**, к описанной выше информационной проблеме следует отнести и лингвистическую проблему, когда не хватает производительности существующих персональных компьютеров для полной (программной) обработки поступающей текстовой информации различного характера.

Задача сегодняшнего этапа – максимальное приближение к практическому внедрению существующих теоретических основ искусственного интеллекта, их дополнение новыми теоретическими разработками с учетом последних достижений инженерной практики с последующим учетом перспективных тенденций и особенностей новых микроэлектронных компонентов вычислительной техники. В связи с изложенным выше актуальность исследований обосновывается необходимостью развития и разработки эффективных формализованных методик архитектурно-структурной организации и проектирования, которые позволят не только проектировать компьютерные системы с обработкой общеязыковых и предметных знаний, но и *автоматизировать процесс* *построения баз знаний предметных областей на основе обработки больших объемов текстовой информации, создавая электронные коллекции баз знаний предметных дисциплин и на их основе проводя сложные междисциплинарные научные исследования.*

## Постановка задачи

Необходимо разработать методологию проектирования и архитектуру программно-аппаратного комплекса средств реализации интегрированной информационной технологии (ИнИТ) автоматизированного построения онтологий предметных областей, названного инструментальным комплексом онтологического назначения (ИКОН), информационную и функциональную модели всех процессов, в совокупности составляющих указанную технологию. Общетеоретические основы проектирования аппаратных средств (или онтолого-управляемых информационных систем с обработкой предметно-ориентированных знаний) описаны в [6].

Модель проектируемой ИнИТ, состоящей из ряда информационных технологий (ИТ) (процессов) и направленной на решение задачи автоматизированного построения онтологий в произвольных (в научно-технической сфере) предметных областях, можно представить следующей системой

$$S = <P, A, X, \Lambda>,$$

(1)

где:

$P = \{p_i\}, i = \overline{1,n}$ – множество процессов, реализующих ИнИТ;

$A = \{A_j\}, j = \overline{1,m}, m \geq n$ – множество алгоритмов, реализующих множество процессов $\{p_i\}$, в совокупности составляющих последовательность информационных технологий и интегрированную информационную технологию в целом. Причем может быть несколько алгоритмов, реализующих некоторый процесс $p_i$;

$X$ – множество сущностей, описывающих заданную ПдО и участвующих в реализации алгоритмов $\{A_j\}$;

$\Lambda$ – обобщенная архитектура программно-аппаратных средств, участвующих в реализации ИнИТ.

Обобщенная архитектура инструментального комплекса ИКОН описывается тройкой

$$\Omega = <\Omega^H, \Lambda, \Omega^{CY}>$$

(2)

и проектируется в соответствии с онтологическим (*O*) методом [6]. Здесь:

$\Omega^H$ – подсистема информационного ресурса;

$\Lambda$ – подсистема программно-аппаратных средств, реализующих ИнИТ;

$\Omega^{CY}$ – подсистема программного уровня управления множеством процессов $\{p_i\}$ в соответствии с *O*-моделью [6].

Блок-схема Ω-архитектуры представлена на рис. 1.

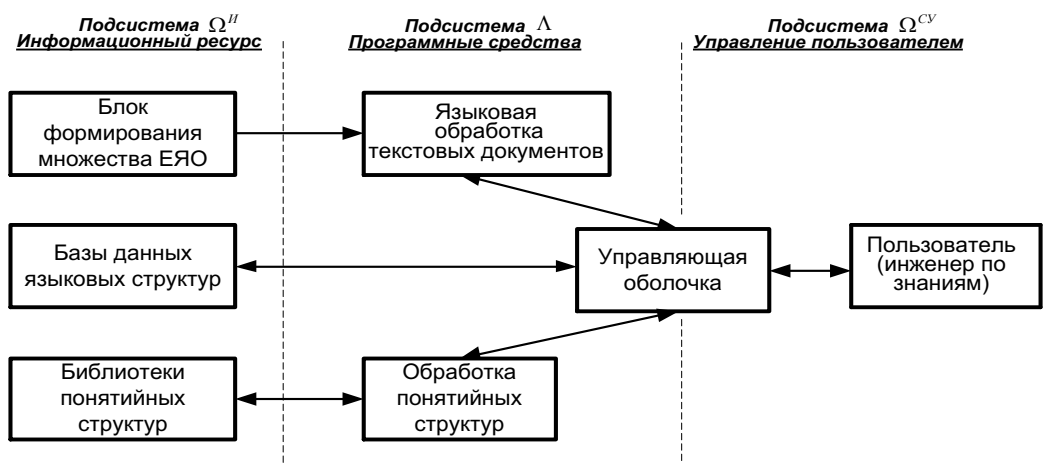

Рис. 1. Блок-схема Ω-архитектуры ИКОН

В соответствии с [2] ИКОН состоит из трех подсистем и представляет собой интеграцию разного рода информационных ресурсов (ИР), программно-аппаратных средств обработки и программного уровня управления основными и подготовительными процессами, которые, взаимодействуя между собой, реализуют совокупность алгоритмов автоматизированного итерационного построения понятийных структур предметных знаний, их накопления и/или системной интеграции.

Подсистема $\Omega^И$ "***Информационный ресурс***" включает блоки формирования лингвистического корпуса текстов, баз данных языковых структур и библиотек понятийных структур. Первый компонент представляет собой различные источники текстовой информации, поступающей на обработку в систему. Второй компонент представляет собой различные базы данных обработки языковых структур, часть из которых формируется (наполняется данными) в процессе обработки текстовых документов (ТД), а другая – формируется до процесса построения онтологии предметной области (О ПдО) и, по сути, является электронной коллекцией (ЭлК) различных толковых словарей. Третий компонент представляет собой совокупность библиотек понятийных структур разного уровня представления (от наборов терминов и понятий до высокоинтегрированной онтологической структуры междисциплинарных знаний) и является результатом реализации некоторого проекта (проектирования онтологии ПдО).

Подсистема $\Lambda$ "***Программно-аппаратные средства***" включает блоки обработки языковых и понятийных структур и управляющую графическую оболочку (УГО). Последняя во взаимодействии с инженером по знаниям, осуществляет общее управление процессом реализации связанных информационных технологий.

Подсистема управления $\Omega^{СУ}$ осуществляет подготовку и реализацию процедур предварительного этапа проектирования, а на протяжении всего процесса осуществляет контроль и проверку результатов выполнения этапов проектирования, принимает решение о степени их завершённости (и в случае необходимости – повторении некоторых из них).

Итак, общая задача является составной и включает следующие подзадачи:
1) разработать функциональную модель интегрированной информационной технологии;
2) разработать информационную модель множества процессов $P$, составляющих ИнИТ;
3) разработать подсистему программных средств $\Lambda$, реализующих множество алгоритмов $A$;
4) при решении общей задачи учитывать критерии эффективности $O$-метода, в частности, уровень автоматизации построения онтологической базы знаний ПдО и ограничение реального времени получения результата.

## Информационная модель ИнИТ

На рис. 2 представлена блок-схема интегрированной информационной технологии (или информационной модели) автоматизированного построения онтологий предметных областей. На ней приняты следующие обозначения:

1 – технология поиска в различных источниках текстовых документов (ТД), релевантных к заданной предметной области (ПдО);
2 – технология автоматического лингвистического анализа ТД, описывающих заданную ПдО. В случае разрешения грамматической омонимии возможен возврат к блоку 1 для поиска соответствующей текстовой информации в лингвистических словарях;
3 – технология извлечения из множества ТД онтолого-информационных структур и их хранение в соответствующих базах данных. В случае разрешения лексической неоднозначности возможен возврат к блоку 2 для построения дополнительных синтактико-семантических структур;
4 – технология формально-логического представления и интеграции онтологических структур в онтологическую базу знаний ПдО;
5 – технология обработки и управления данными и знаниями большого объема;
6 – источники текстовых документов (сеть Интернет, монографии, научно-технические статьи, электронные коллекции ТД и др.);
7 – лингвистический корпус текстов (ЛКТ), описывающий заданную ПдО;
8 – синтаксические и поверхностно-семантические структуры ТД;
9 – множество онтолого-информационных структур как результат обработки ТД;
10 – онтологическая база знаний заданной ПдО, множества терминов и понятий ПдО;
11 – библиотеки онтологий предметных областей, базы данных терминов, понятий и ТД предметных областей.

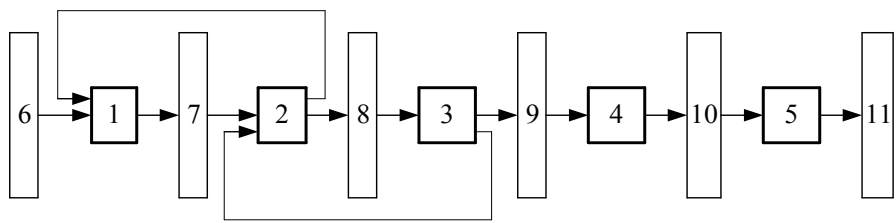

Рис. 2. Блок-схема интегрированной информационной технологии

Далее, в процессе разработки ИнИТ выполняется декомпозиция общей информационной модели на ее составляющие информационные модели, соответствующие отдельным ИТ, т. е. формируется иерархическая структура в соответствии с методикой проектирования онтологии процессов [3].

В качестве примера рассмотрим информационную модель блока 2 (рис. 2) – технологии автоматического лингвистического анализа ЕЯО, описывающих заданную ПдО.

Архитектура современных знание-ориентированных информационных систем (ЗОИС) с естественно-языковым представлением и обработкой знаний включает онтологическую составляющую, которую в общем виде можно интерпретировать как концептуальную базу знаний. Такая база знаний представляется в виде ориентированного графа, вершинами которого являются объекты, описывающие концепты, а дугами – множество концептуальных отношений, связывающих между собой концепты. Другой важной особенностью указанной архитектуры является разделение и отдельная обработка семантики первой и второй ступени [7], что в общем случае означает разделение внутриязыкового и внеязыкового процессинга [8] и переход к формально-логическому представлению исходного текста.

Указанные особенности архитектуры современных ЗОИС трансформируют традиционную модель (и соответствующую информационную технологию) обработки ЕЯО в формальную модель следующего вида

$$F = <T, W, SS^1, O, S^2, I>, \qquad (3)$$

где $T$ – множество обрабатываемых ЕЯТ, составляющих лингвистический корпус текстов;

$W$ – множество словоформ, входящих в $T$;

$SS^1$ – множество синтактико-семантических структур первой ступени, описывающих $T$;

$O$ – множество онтологических структур, отображающих множества $W$ и $SS^1$ в $S^2$;

$S^2$ – множество семантических структур второй ступени, описывающих множество процессов, зафиксированных в $T$;

$I$ – множество информационно-кодовых представлений $S^2$.

Опишем объекты формальной модели.

Множество $T$ представляет совокупность естественно-языковых текстов, характеризующихся стилями делового и научно-технического характера.

Цепочка $W \to SS^1$ в классическом понимании представляет грамматический анализ ЕЯТ. В отличие от традиционных линейного и сильнокодированного методов анализа здесь использован смешанный метод анализа. Суть его состоит в том, что в лексикографической базе данных полное множество $W$ представлено в таблицах двух типов: таблицами лексем с соответствующими морфологическими, синтаксическими и семантическими характеристиками и таблицами флексий для всех полнозначных, изменяющихся частей речи. При этом алгоритмы формирования парадигмы лексем просты: в таблицах лексики указаны основы лексем и соответствующие коды для выбора записей из таблиц флексий. Нефлексийные изменения учитываются соответствующими алгоритмами анализа согласования и управления синтаксических связей. Описанная структура грамматического анализа однозначно соответствует эффективному отображению функциональных операторов на программно-аппаратный уровень реализации.

Множество $O$ онтологических структур в идеале представляет языково-онтологическую картину мира [7, 9].

Множество $SS^1$ формируется и интерпретируется итерационно программой синтактико-семантического анализа. Основной операцией этого анализа является распознавание синтаксических и семантических отношений, связывающих слова текста. Разрешение семантической неоднозначности осуществляется путём обращения к множеству онтологических структур $O$. На основе построенных деревьев разбора фраз строится категориальная сеть, представляющая собой семантическое пространство $S^2$ текста. В качестве компьютерного представления такого пространства текста удобно использовать, в частности, онтологический граф информационно-кодовых представлений $I$.

Цепочки преобразования информации $T \to W \to SS^1$ и $O \to S^2 \to I$, по сути, представляют (соответственно) базовые процедуры анализа и понимания ЕЯТ, средствами интерпретации которых являются программные средства блока 2 и (частично) блока 3 на рис. 2.

## Функциональная модель ИнИТ

Для проектирования функциональных моделей (функционального моделирования) программного обеспечения и различного рода информационных технологий известно ряд общепринятых стандартных методологий и языков функционального моделирования, таких как *IDEF, DFD, UML*.

*IDEF* – методологии семейства ICAM (Integrated Computer-Aided Manufacturing) для решения задач моделирования сложных систем, позволяют отображать и анализировать функциональные модели сложных систем в различных аспектах. При этом широта и глубина обследования процессов в системе определяется самим разработчиком, что позволяет не перегружать создаваемую модель излишними данными. Набор стандартов *IDEF* промышленного моделирования включает ряд стандартов, таких как: *IDEF0, IDEF1x, IDEF3* и других, основанных на графических нотациях, и применяемых для разработки, поддержки больших и сложных промышленных инженерных проектов [10]. По отдельности эти нотации предназначены для создания различных моделей, таких как "функциональная модель", "информационная модель" и "модель процессов".

*Метод функционального моделирования IDEF0* предназначен для формализации и описания процессов, в котором рассматриваются логические отношения между объектами. В соответствии со стандартом IDEF0 описание выглядит как общий процесс с входами, выходами, управлением и механизмом, который постепенно детализируется до необходимого уровня. Для описания того, какой смысл вкладывается в названия функциональных блоков и стрелок используются словари описания [11, 12, 13].

В настоящее время диаграммы структурного системного анализа IDEF-SADT продолжают использоваться для построения и детального анализа функциональной модели существующих на предприятии бизнес-процессов, а также для разработки новых бизнес-процессов. Основной недостаток данной методологии связан с отсутствием явных средств для объектно-ориентированного представления моделей сложных систем.

Основой *DFD*-методологии является графическое моделирование информационных систем путем построения диаграмм потоков данных.

Модель системы в контексте DFD представляется в виде некоторой информационной структуры, основными компонентами которой являются различные потоки данных, которые переносят информацию от одной подсистемы к другой. Каждая из подсистем выполняет определенные преобразования входного потока данных и передает результаты обработки информации в виде потоков данных для других подсистем.

Информационная модель системы строится как некоторая иерархическая схема, в основе которой находится так называемая *контекстная диаграмма*. На диаграмме декомпозиции исходная модель последовательно представляется в виде моделей подсистем соответствующих процессов преобразования данных.

Основной недостаток этой методологии также связан с отсутствием явных средств для объектно-ориентированного представления моделей сложных систем, а также для представления сложных алгоритмов обработки данных [14]. Поскольку на диаграммах DFD не указываются характеристики времени выполнения отдельных процессов и передачи данных между процессами, то модели систем, реализующих синхронную обработку данных, не могут быть адекватно представлены в нотации DFD. Все эти особенности ограничили возможности ее широкого применения и послужили основой для включения соответствующих средств в унифицированный язык моделирования.

**UML** (англ. Unified Modeling Language – унифицированный язык моделирования) – язык графического описания для объектного моделирования в области разработки программного обеспечения [15]. UML является широко используемым стандартом де-факто объектно-ориентированного визуального языка моделирования. Он разрабатывался с учетом преимуществ и недостатков SADT (IDEF) и DFD методов. В UML есть три основных вида моделей [16, 17]:
- статическая модель (static model);
- динамическая модель (dynamic model).
- физическая модель (physical model)

Статическая модель описывает элементы системы и их взаимоотношения (классы, атрибуты, операторы). Одной из реализаций статической модели является диаграмма классов (class diagram). Динамическая модель описывает поведение системы, например, изменение программных сущностей (software entities) во время выполнения приложения. К динамическим моделям относятся: диаграмма прецедентов, также: вариантов использования, сценариев использования (use case diagram); диаграмма активности, также деятельности (activity diagram). Физическая модель отображает неизменяемую структуру программных сущностей, в частности файлов исходного кода, библиотек, исполняемых файлов, и отношений между ними.

В настоящее время разработаны средства визуального программирования на основе UML, обеспечивающие интеграцию, включая прямую и обратную генерацию кода программ, с наиболее распространенными языками и средами программирования, такими как Java, C++, C#, Visual Basic, Object Pascal/Delphi. Поскольку при разработке языка UML были приняты во внимание многие передовые идеи и методы, можно ожидать, что на очередные версии языка UML также окажут влияние и другие перспективные технологии и концепции. Кроме того, на основе языка UML могут быть определены многие новые перспективные методы. Язык UML может быть расширен без переопределения его ядра.

UML технологии стали основой для разработки и реализации во многих инструментальных средствах: в средствах визуального и имитационного моделирования, а также в CASE-средствах самого различного целевого назначения. Более того, заложенные в языке UML потенциальные возможности могут быть использованы не только для объектно-ориентированного моделирования систем, но и для представления знаний в интеллектуальных системах, которыми, по существу, являются перспективные сложные программно-технологические комплексы. Принимая во внимание все преимущества языка UML, на его основе была спроектирована функциональная модель ИнИТ.

*Функциональная модель ИнИТ представляет собой набор диаграмм трёх видов*:
- диаграмма вариантов использования (рис. 3);
- диаграмма активности (рис. 4);
- диаграмма классов (рис. 5).

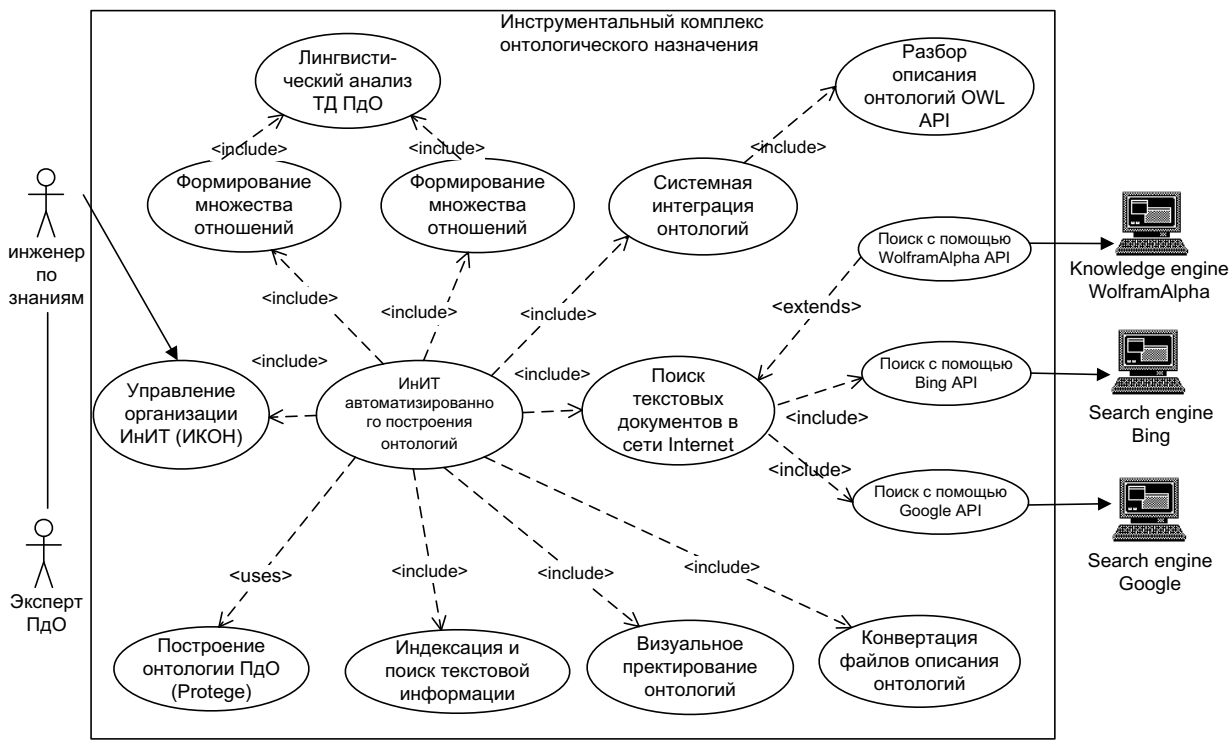

Рис. 3. UML-диаграмма вариантов использования ИнИТ (ИКОН)

Целями разработки *диаграммы вариантов использования* являются:
1. Определить общие границы и контекст моделируемой предметной области на начальных этапах проектирования системы.
2. Сформулировать общие требования к функциональному поведению проектируемой системы.
3. Разработать исходную концептуальную модель системы для её последующей детализации в форме логических и физических моделей.
4. Подготовить исходную документацию для взаимодействия разработчиков системы с её заказчиками и пользователями.

Суть данной диаграммы состоит в следующем: проектируемая система представляется в виде множества сущностей или акторов (actor), взаимодействующих с системой с помощью так называемых вариантов использования. При этом актором или действующим лицом называется любая сущность, взаимодействующая с системой извне. Это может быть человек, техническое устройство, программа или любая другая система, которая может служить источником воздействия на моделируемую систему. В свою очередь, вариант использования служит для описания сервисов, которые система предоставляет актору.

С помощью *диаграммы активности* (рис. 4) можно изучать поведение системы с использованием моделей потока данных и потока управления. Диаграмма активности отображает некоторый алгоритм, описывающий жизненный цикл объекта, состояния которого могут меняться.

Диаграмма активности отличается от блок-схемы, которая описывает лишь шаги алгоритма. Диаграмма активности имеет более широкую нотацию. Например, на ней можно указывать состояния объектов.

*Диаграмма классов* описывает структуру объектов ИнИТ: их индивидуальность, отношения с другими объектами, атрибуты, функции и процедуры. Модель классов создает контекст для диаграмм состояний и взаимодействия. На рис. 5 представлен фрагмент UML-диаграмма классов ИнИТ:
– класс **ArchiveSearch** реализует технологию обработки и управления хранилищами данных и знаний;
– класс **GraphEditor** реализует технологию формально-логического представления и интеграции онтологических структур;
– класс **ControlGraphicShell** реализует управляющую графическую оболочку, которая осуществляет общее управление процессом реализации связанных информационных технологий;
– интерфейс для доступа к внешней библиотеке **KonspektLib**, реализующей функции лингвистического анализа.

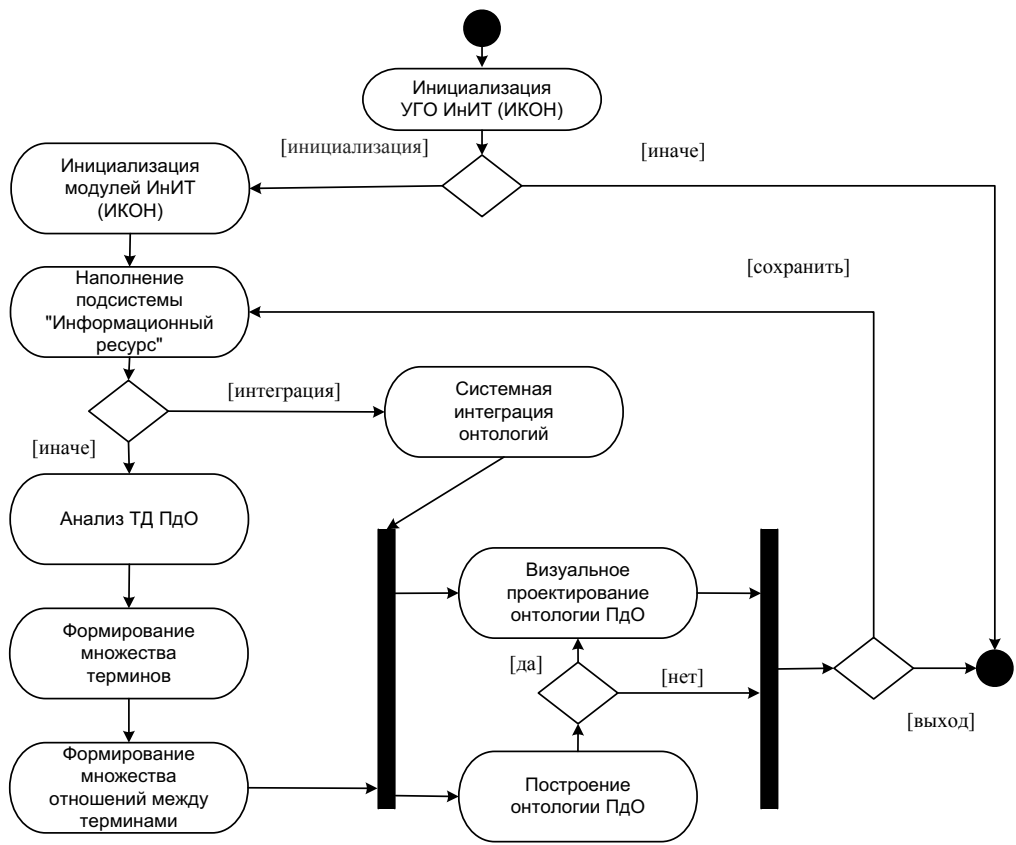

Рис. 4. UML-диаграмма активности ИнИТ (ИКОН)

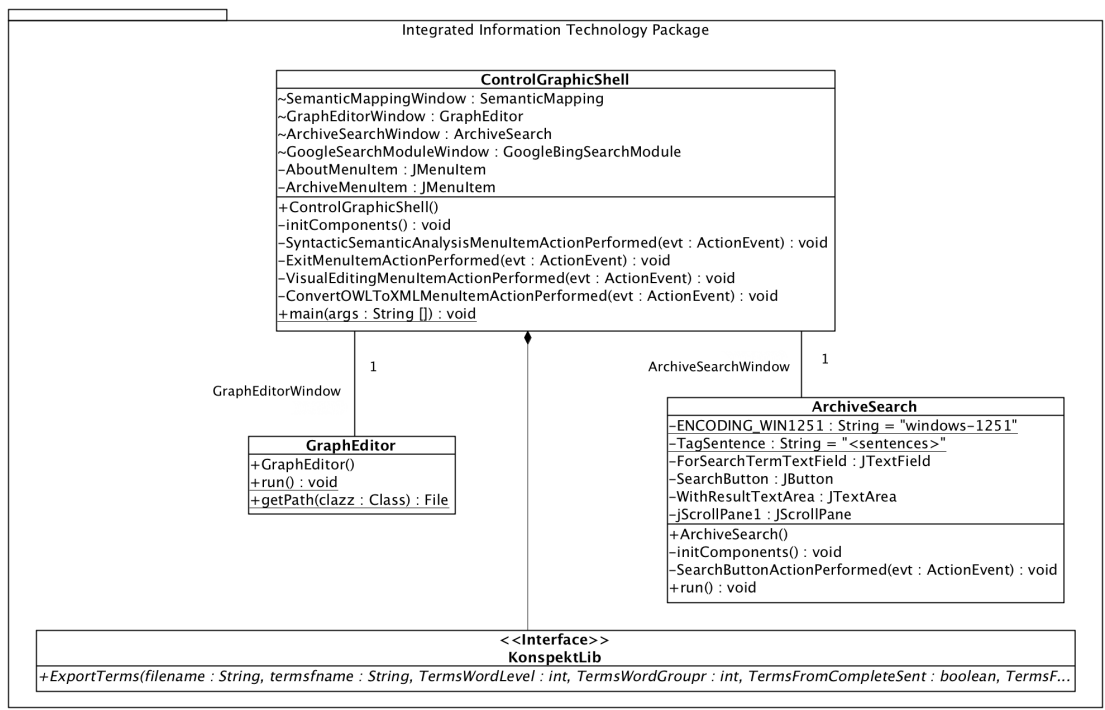

Рис. 5. Фрагмент UML-диаграммы классов ИнИТ

## Подсистема программных средств

Опишем некоторые программные модули, входящие в подсистемы программных средств и программного уровня управления множеством процессов, реализующих ИнИТ.

Программный модуль "*Управляющая графическая оболочка*" [18].

В состав программно-аппаратных средств ИнИТ (ИКОН) входит Управляющая графическая оболочка (УГО), в которую включены процедуры и функции управления множеством процессов, реализующих ИнИТ (ИКОН).

*Графический интерфейс УГО.*

Проектирование графического интерфейса УГО осуществлялось в соответствии с общепринятыми рекомендациями построения графического интерфейса пользователя (ГПИ) [19, 20]. Рассмотрим основные элементы главного окна УГО и их назначение (рис. 6).

Меню запуска прикладных программ и модулей, которые реализуют информационные технологии проектирования онтологий ПдО и системной интеграции междисциплинарных знаний. Доступны следующие модули и программы: модуль лингвистического анализа естественно-языковых текстов; модуль визуального проектирования онтологий, модуль конвертации OWL (Ontology Web Language); модуль системной интеграции онтологий; модуль системы поиска; модуль архива терминов; вызов редактора Protégé для построения и редактирования онтологий.

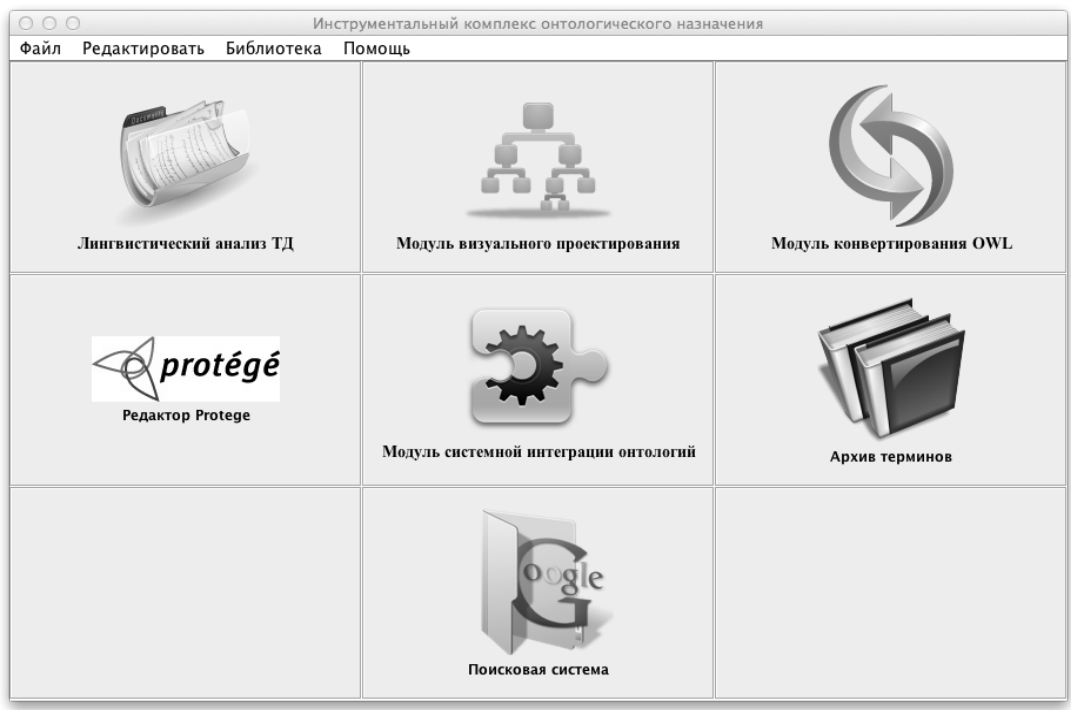

Рис. 6. Элементы ГПИ главного окна УГО ИКОН.

Структурная схема УГО ИнИТ (ИКОН) представлена на рис. 7. На нём приняты такие сокращения: GUI – Graphical User Interface, XML – Extensible Markup Language, МВП – модуль визуального проектирования. УГО взаимодействует с подсистемой Естественный Интеллект (ЕИ), осуществляет общее управление процессом реализации связанных информационных технологий.

УГО выполняет следующие функции:
- во взаимодействии с инженером по знаниям осуществляет предварительное наполнение среды материалами электронных коллекций энциклопедических, толковых словарей и тезаурусов, описывающих домен предметных знаний;
- обеспечивает запуск и последовательность исполнения прикладных программ, реализующих составные информационные технологии проектирования онтологии ПдО и системной интеграции междисциплинарных знаний (примером составной технологии является автоматизированное построение тезаурусов ПдО для поисковой системы);
- отображает ход процесса проектирования;
- содержит позиции меню для запуска, как последовательностей, так и отдельных прикладных программ, используемых в процессе проектирования;
- обеспечивает интерфейс с подсистемой ЕИ;
- индицирует сообщения о текущем состоянии проекта, его наполнении информационными ресурсами;
- обеспечивает обмен информацией между прикладными программами и базами данных.

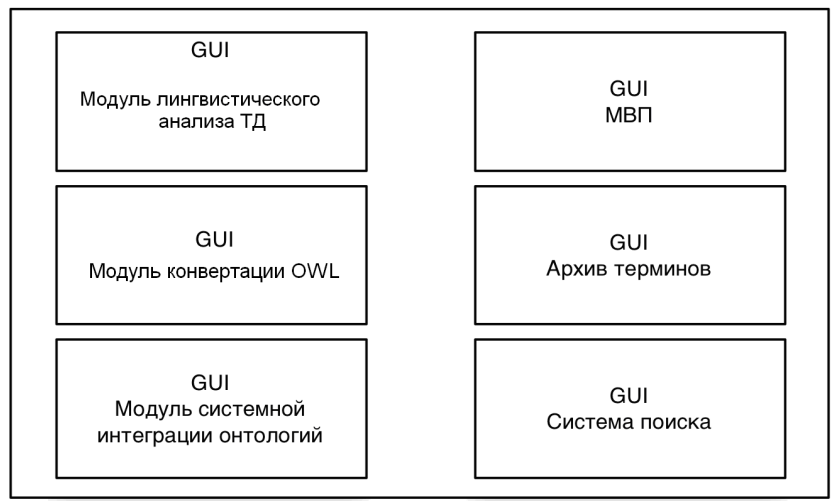

Рис. 7. Структурная схема УГО ИнИТ (ИКОН)

Программная реализация ИКОН была выполнена с использованием концепции **Model-View-Controller** (MVC) – "Модель-Представление-Контроллер" – основная цель применения этой концепции состоит в разделении бизнес-логики (*модели*) от её визуализации (*представления*, *вида*) [21, 22].

MVC – схема использования нескольких шаблонов проектирования с помощью которых модель данных приложения, пользовательский интерфейс и взаимодействие с пользователем разделены на три отдельных компонента так, что модификация одного из компонентов оказывает минимальное воздействие на остальные.

Концепция *MVC* позволяет разделить данные, представление и обработку действий пользователя на три отдельных компонента:
- *Модель* (*Model*). Модель предоставляет знания: данные и методы работы с этими данными, реагирует на запросы, изменяя своё состояние. Она не содержит информации, как эти знания можно визуализировать.
- *Представление*, *вид* (*View*). Отвечает за отображение информации (визуализация). Часто в качестве представления выступает форма (окно) с графическими элементами.
- *Контроллер* (*Controller*). Обеспечивает связь между пользователем и системой: контролирует ввод данных пользователем и использует модель и представление для реализации необходимой реакции.

Важно отметить, что как *представление*, так и *контроллер* зависят от *модели*. Однако *модель* не зависит ни от *представления*, ни от *контроллера*. Тем самым достигается назначение такого разделения: оно позволяет строить *модель* независимо от *визуального представления*, а также создавать несколько различных *представлений* для одной *модели*. На рис. 8 представлена реализация концепции MVC в модуле поиска ТД в Internet ИнИТ (ИКОН).

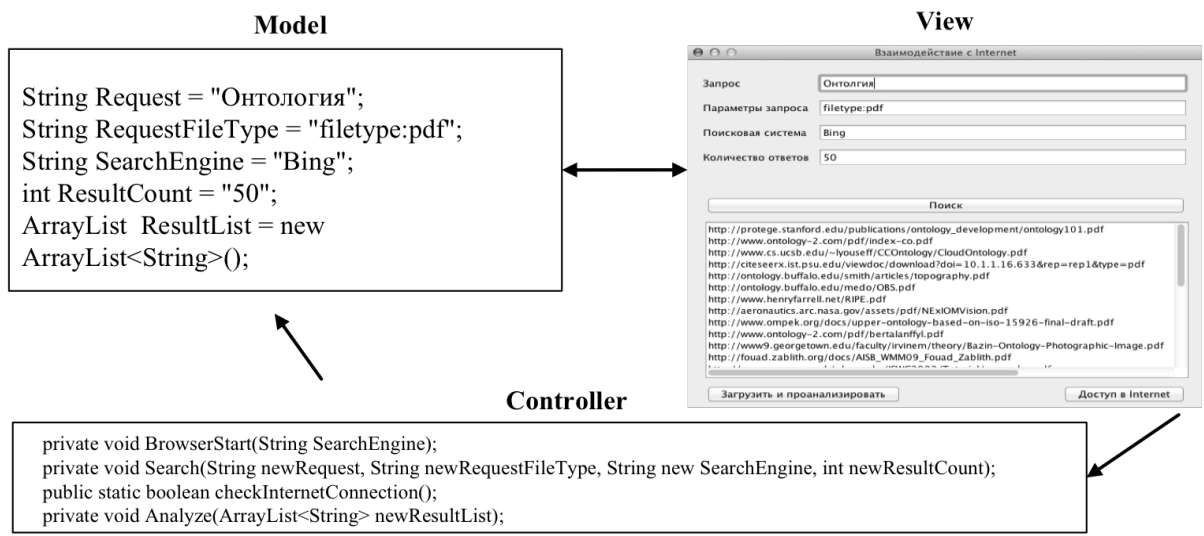

Рис. 8 Реализация концепции MVC в модуле поиска ТД в Internet ИнИТ (ИКОН).

Перечислим этапы реализации концепции MVC в модуле поиска ТД в Internet ИнИТ (ИКОН). Объекты *модель*, *вид* и *контроллер* инициализированы, тогда:
1. *Вид* (View) регистрируется в качестве слушателя (listener) на *модели* (Model). Любые изменения в исходных данных *модели* немедленно приведут к уведомлению (notification) на изменение, которое получает *вид*. Еще раз следует обратить внимание на то, что *модель* независима от *вида* и *контроллера* (Controller).

2. *Контроллер* связан с *видом*. Это означает, что любые действия (actions), которые выполняются на представлении (*виде*), будут вызывать зарегистрированный метод *слушателя* в классе *контроллера*.
3. *Контроллер* получает ссылку на базовую *модель*.
4. *Вид* распознаёт взаимодействие пользователя с ГПИ, например, нажатие на кнопку. Это действие определяет вызов метода *слушателя*.
5. *Вид* вызывает соответствующий метод *контроллера*.
6. *Контроллер* осуществляет доступ к *модели*, возможно, обновляя её с учётом действий пользователя.
7. Если *модель* была изменена, она уведомляет *вид*, об изменениях.

Программная реализация ИКОН выполнена с использованием платформы Java Swing Framework. В отличие от других платформ, она не только предоставляет интерфейс разработки на основе шаблона MVC, но и сама реализована на его основе. Представлением является класс – наследник класса Frame. Вследствие организации событийной модели Java на интерфейсах, контроллер представляет собой набор анонимных классов обработки соответствующих событий. Как и остальные платформы, Swing предоставляет разработку модели программисту.

## Выводы

В работе предложена методология проектирования и архитектура программных средств реализации ИнИТ автоматизированного построения онтологий предметных областей. Разработаны информационная и функциональная модели всех процессов, в совокупности составляющих ИнИТ. Приведено подробное описание подсистемы программных средств ИКОН, в частности УГО. Проанализированы технологии моделирования (IDEF, DFD, UML) и проектирования (MVC) сложных систем. Приведены UML-диаграммы, описывающие функциональную модель ИнИТ (ИКОН). За рамками разработки ИКОН остались проблемы восприятия информации, представленной в аналитической либо графической форме. Они определяют содержание последующих этапов научных исследований, равно как и анализ эффективности применения комплекса при решении конкретных прикладных задач.

## Литература